\documentclass[11pt,letterpaper]{article}
\usepackage{naaclhlt2015}
\usepackage{times}
\usepackage{latexsym}
\usepackage{amsmath}
\usepackage{enumerate}
\usepackage{graphicx}
\usepackage{subfigure}
\usepackage{multirow}

\title{Robustly Leveraging Prior Knowledge in Text Classification}
\author{Biao Liu\\
	    Tsinghua University\\
	    Haidian District, Beijing, China\\
	    {\tt liubiao2638@gmail.com}
	  \And
	Minlie Huang\\
  	Tsinghua University\\
  	Haidian District, Beijing, China\\
  {\tt aihuang@tsinghua.edu.cn}}

\date{}

\begin{document}
\maketitle

\begin{abstract}

Prior knowledge has been shown very useful to address many natural language processing tasks. Many approaches have been proposed to formalise a variety of knowledge, however, whether the proposed approach is robust or sensitive to the knowledge supplied to the model has rarely been discussed. In this paper, we propose three regularization terms on top of generalized expectation criteria, and conduct extensive experiments to justify the robustness of the proposed methods. Experimental results demonstrate that our proposed methods obtain remarkable improvements and are much more robust than baselines.

\end{abstract}

\section{Introduction}
\label{intro_sec}
We posses a wealth of prior knowledge about many natural language processing tasks.
For example, in text categorization, we know that words such as \textit{NBA, player,} and \textit{basketball} are strong indicators of the \textit{sports} category \cite{ge-fl}, and words like \textit{terrible, boring}, and \textit{messing} indicate a \textit{negative} polarity while words like \textit{perfect, exciting}, and \textit{moving} suggest a \textit{positive} polarity in sentiment classification.

A key problem arisen here, is how to leverage such knowledge to guide the learning process, an interesting problem for both NLP and machine learning communities.
Previous studies addressing the problem fall into several lines.
First,  to leverage prior knowledge to label data \cite{haghighi2006prototype,raghavan2007interactive}.
Second, to encode prior knowledge with a prior on parameters, which can be commonly seen in many Bayesian approaches \cite{andrzejewski2009latent,andrzejewski2011framework}.
Third, to formalise prior knowledge with additional variables and dependencies \cite{li2010sentiment}.
Last, to use prior knowledge to control the distributions over latent output variables \cite{pr,gec,guiding},
which makes the output variables easily interpretable.

However, a crucial problem, which has rarely been addressed, is the bias in the prior knowledge that we supply to the learning model. Would the model be robust or sensitive to the prior knowledge?
 Or, which kind of knowledge is appropriate for the task? Let's see an example: we may be a {\em baseball} fan but unfamiliar with {\em hockey} so that we can provide a few number of feature words of {\em baseball}, but much less of {\em hockey} for a {\em baseball-hockey} classification task.
 Such prior knowledge may mislead the model with heavy bias to \textit{baseball}.
 If the model cannot handle this situation appropriately, the performance
 may be undesirable.

In this paper, we investigate into the problem in the framework of Generalized Expectation Criteria \cite{gec}.
The study aims to reveal the factors of reducing the sensibility of the prior knowledge and therefore to make the model more robust and practical.
To this end,  we introduce auxiliary regularization terms in which our prior knowledge is formalized as distribution over output variables.
Recall the example just mentioned, though we do not have enough knowledge to provide features for class \textit{hockey},
it is easy for us to provide some neutral words, namely words that are not strong indicators of any class, like {\em player} here.
As one of the factors revealed in this paper, supplying neutral feature words can boost the performance remarkably, making the model more robust.

More attractively, we do not need manual annotation to label these neutral feature words in our proposed approach.

More specifically, we explore three regularization terms to address the problem: (1) a regularization term associated with neutral features;
(2) the maximum entropy of class distribution regularization term; and (3) the KL divergence between reference and predicted class distribution.
 For the first manner, we simply use the most common features as neutral features and assume the neutral features are distributed uniformly over class labels.
For the second and third one, we assume we have some knowledge about the class distribution which will be detailed soon later.

To summarize, the main contributions of this work are as follows:
\begin{itemize}

\item We explore three regularization terms to make the model more robust: a regularization term associated with neutral features, the maximum entropy of class distribution regularization term, and the KL divergence between reference and predicted class distribution.

\item Experiments demonstrate that the proposed approaches outperform baselines and work much more robustly. 

\end{itemize}

The rest of the paper is structured as follows: In Section 2, we briefly describe the generalized expectation criteria and present the proposed regularization terms. In Section 3, we conduct extensive experiments to justify the proposed methods. We survey related work in Section 4, and summarize our work in Section 5.
\section{Method}
We address the robustness problem on top of GE-FL \cite{ge-fl}, a GE method which leverages labeled features as prior knowledge. A labeled feature is a strong indicator of a specific class and is manually provided to the classifier. For example, words like {\em amazing, exciting} can be labeled features for class {\em positive} in sentiment classification.

\subsection{Generalized Expectation Criteria}
Generalized expectation (GE) criteria \cite{gec} provides us a natural way to directly constrain the model in the preferred direction. For example, when we know the proportion of each class of the dataset in a classification task, we can guide the model to predict out a pre-specified class distribution.

Formally, in a parameter estimation objective function, a GE term expresses preferences on the value of some constraint functions about the model's expectation. Given a constraint function $G({\rm x}, y)$, a conditional model distribution $p_\theta(y|\rm x)$, an empirical distribution $\tilde{p}({\rm x})$ over input samples and a score function $S$, a GE term can be expressed as follows:
\begin{equation}
\label{eq-gec}
S(E_{\tilde{p}({\rm x})}[E_{p_\theta(y|{\rm x})}[G({\rm x}, y)]])
\end{equation}

\subsection{Learning from Labeled Features}
Druck et al. \shortcite{ge-fl} proposed GE-FL to learn from labeled features using generalized expectation criteria. When given a set of labeled features $K$,  the reference distribution over classes of these features is denoted by $\hat{p}(y| x_k), k \in K$. GE-FL introduces the divergence between this reference distribution and the model predicted distribution $p_\theta(y | x_k)$ , as a term of the objective function:
\begin{equation}
\label{eq-ge-fl}
\mathcal{O} = \sum_{k \in K} KL(\hat{p}(y|x_k) || p_\theta(y | x_k)) + \sum_{y,i} \frac{\theta_{yi}^2}{2 \sigma^2}
\end{equation}
where $\theta_{yi}$ is the model parameter which indicates the importance of word $i$ to class $y$. The predicted distribution $p_\theta(y | x_k)$ can be expressed as follows:
\[
p_\theta(y | x_k) = \frac{1}{C_k} \sum_{\rm x} p_\theta(y|{\rm x})I(x_k)
\]
in which $I(x_k)$ is 1 if feature $k$ occurs in instance ${\rm x}$ and 0 otherwise, $C_k = \sum_{\rm x} I(x_k)$ is the number of instances with a non-zero value of feature $k$, and $p_\theta(y|{\rm x})$ takes a softmax form as follows:
\[
p_\theta(y|{\rm x}) = \frac{1}{Z(\rm x)}\exp(\sum_i \theta_{yi}x_i).
\]

To solve the optimization problem, L-BFGS can be used for parameter estimation.

In the framework of GE, this term can be obtained by setting the constraint function $G({\rm x}, y) = \frac{1}{C_k} \vec I (y)I(x_k)$, where $\vec I(y)$ is an indicator vector with $1$ at the index corresponding to label $y$ and $0$ elsewhere.

\subsection{Regularization Terms}
GE-FL reduces the heavy load of instance annotation and performs well when we provide prior knowledge with no bias. In our experiments, we observe that comparable numbers of labeled features for each class have to be supplied. But as mentioned before, it is often the case that we are not able to provide enough knowledge for some of the classes. For the {\em baseball-hockey} classification task, as shown before, GE-FL will predict most of the instances as {\em baseball}.
In this section, we will show three terms to make the model more robust.

\subsubsection{Regularization Associated with Neutral Features}
Neutral features are features that are not informative indicator of any classes, for instance, word {\em player} to the {\em baseball-hockey} classification task.
Such features are usually frequent words across all categories. When we set the preference distribution of the neutral features to be uniform distributed, these neutral features will prevent the model from biasing to the class that has a dominate number of labeled features.

Formally, given a set of neutral features $K^{'}$, the uniform distribution is $\hat{p}_u(y|x_k) = \frac{1}{|C|}, k \in K^{'}$, where $|C|$ is the number of classes. The objective function with the new term becomes
\begin{equation}
\label{eq-nf}
\mathcal{O}_{NE} = \mathcal{O} + \sum_{k \in K^{'}} KL(\hat{p}_u(y|x_k) || p_\theta (y | x_k)).
\end{equation}
Note that we do not need manual annotation to provide neutral features.
One simple way is to take the most common features as neutral features.
Experimental results show that this strategy works successfully.

\subsubsection{Regularization with Maximum Entropy Principle}
\label{max-ent}

Another way to prevent the model from drifting from the desired direction is to constrain the predicted class distribution on unlabeled data.
When lacking knowledge about the class distribution of the data, one feasible way is to take maximum entropy principle, as below:
\begin{equation}
\label{eq-me}
\mathcal{O}_{ME} = \mathcal{O} + \lambda \sum_{y} p(y) \log p(y)
\end{equation}
where $p(y)$ is the predicted class distribution, given by
$
p(y) = \frac{1}{|X|} \sum_{\rm x} p_\theta(y | \rm x).
$
To control the influence of this term on the overall objective function, we can tune $\lambda$ according to the difference in the number of labeled features of each class. In this paper, we simply set $\lambda$ to be proportional to the total number of labeled features, say $\lambda = \beta |K|$. 

This maximum entropy term can be derived by setting the constraint function to $G({\rm x}, y) = \vec I(y)$. Therefore, $E_{p_\theta(y|{\rm x})}[G({\rm x}, y)]$ is just the model distribution $p_\theta(y|{\rm x})$ and its expectation with the empirical distribution $\tilde{p}(\rm x)$ is simply the average over input samples, namely $p(y)$. When $S$ takes the maximum entropy form, we can derive the objective function as above.

\subsubsection{Regularization with KL Divergence}
Sometimes, we have already had much knowledge about the corpus, and can estimate the class distribution roughly without labeling instances.
Therefore, we introduce the KL divergence between the predicted and reference class distributions into the objective function.

Given the preference class distribution $\hat{p}(y)$, we modify the objective function as follows:
\begin{align}
\mathcal{O}_{KL} &= \mathcal{O} + \lambda KL(\hat{p}(y) || p(y))
\end{align}

Similarly, we set $\lambda = \beta |K|$.

This divergence term can be derived by setting the constraint function to $G({\rm x}, y) = \vec I(y)$ and setting the score function to $S(\hat{p}, p) = \sum_i \hat{p}_i \log \frac{\hat{p}_i}{p_i}$, where $p$ and $\hat{p}$ are distributions. Note that this regularization term involves the reference class distribution which will be discussed later.

\section{Experiments}
In this section, 
we first justify the approach when there exists unbalance in the number of labeled features or in class distribution. 
Then, to test the influence of $\lambda$, we conduct some experiments with the method which incorporates the KL divergence of class distribution.
Last, we evaluate our approaches in 9 commonly used text classification datasets.
We set $\lambda = 5|K|$ by default in all experiments unless there is explicit declaration.

The baseline we choose here is GE-FL \cite{ge-fl}, a method based on generalization expectation criteria.

\subsection{Data Preparation}
We evaluate our methods on several commonly used datasets whose themes range from sentiment, web-page, science to medical and healthcare.
We use bag-of-words feature and remove stopwords in the preprocess stage. Though we have labels of all documents, we do not use them during the learning process, instead, we use the label of features.

The {\em movie} dataset, in which the task is to classify the movie reviews as {\em positive} or {\em negtive}, is used for testing the proposed approaches with unbalanced labeled features, unbalanced datasets or different $\lambda$ parameters\footnote{We also experimented on other datasets, and observed similar results.}. All unbalanced datasets are constructed based on the {\em movie} dataset by randomly removing documents of the {\em positive} class.
For each experiment, we conduct 10-fold cross validation.


\subsubsection*{Labeled Features}
As described in \cite{ge-fl}, there are two ways to obtain labeled features. The first way is to use information gain. We first calculate the mutual information of all features according to the labels of the documents and  select the top 20 as labeled features for each class as a feature pool.
Note that using information gain requires the document label, but this is only to simulate
how we human provide prior knowledge to the model.
The second way is to use LDA \cite{lda} to select features.
We use the same selection process as \cite{ge-fl}, where they first train a LDA on the dataset, and then select the most probable features of each topic (sorted by $P(w_i|t_j)$, the probability of word $w_i$ given topic $t_j$).

Similar to \cite{boosting,ge-fl}, we estimate the reference distribution of the labeled features using a heuristic strategy. If there are $|C|$ classes in total, and $n$ classes are associated with a feature $k$, the probability that feature $k$ is related with any one of the $n$ classes is $\frac{0.9}{n}$ and with any other class is $\frac{0.1}{|C| - n}$\footnote{Previous work shows the model is insensitive to the setting of the reference distribution.}.

Neutral features are the most frequent words after removing stop words, and their reference distributions are uniformly distributed. We use the top 10 frequent words as neutral features in all experiments.

\begin{figure*}[!ht]
\setcounter{subfigure}{0}
\centering
	\subfigure[balanced dataset] {
		
		\includegraphics[width=190px]{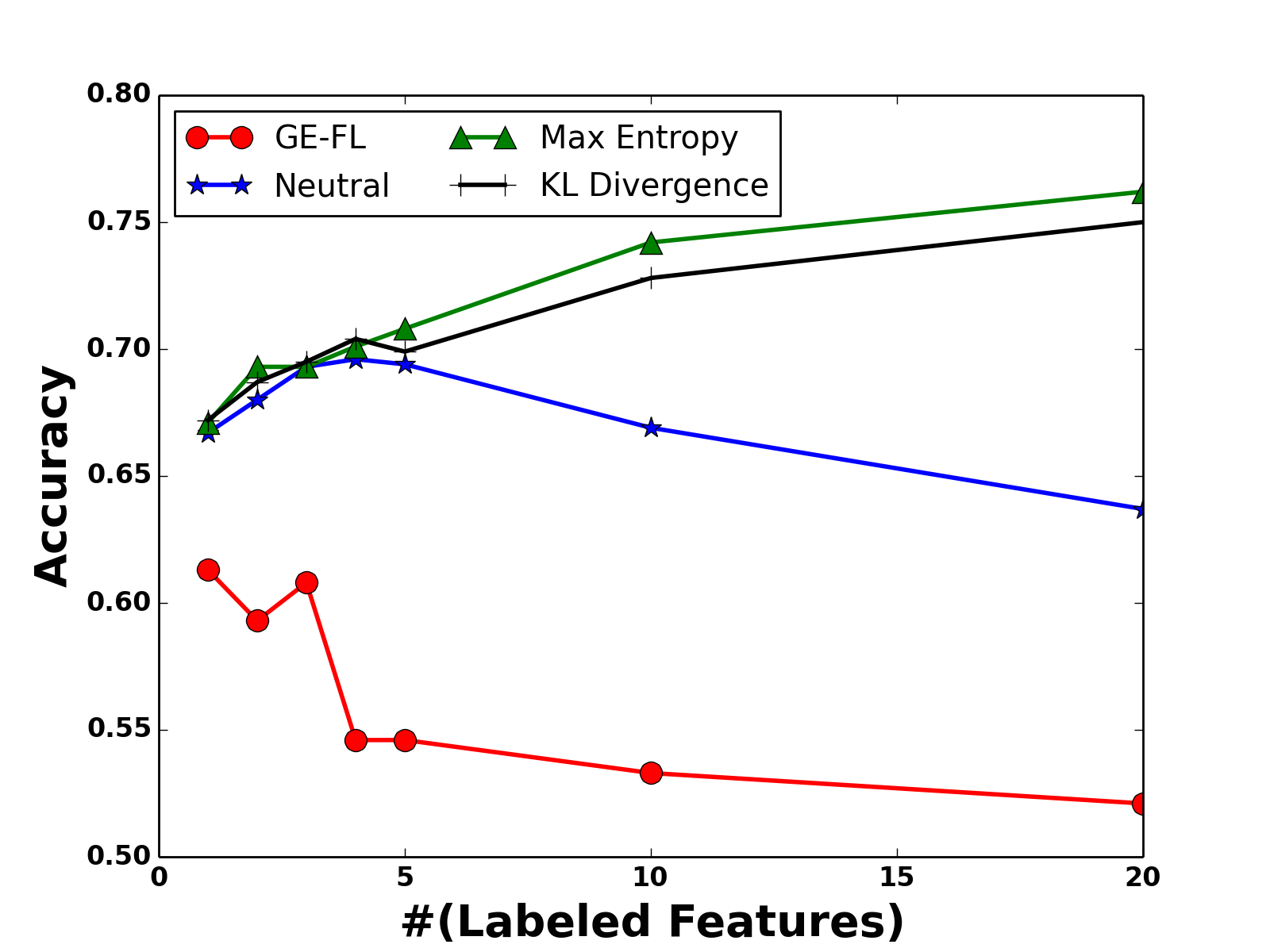}
	}
	\subfigure[unbalanced dataset(1:4)] {
		\includegraphics[width=190px]{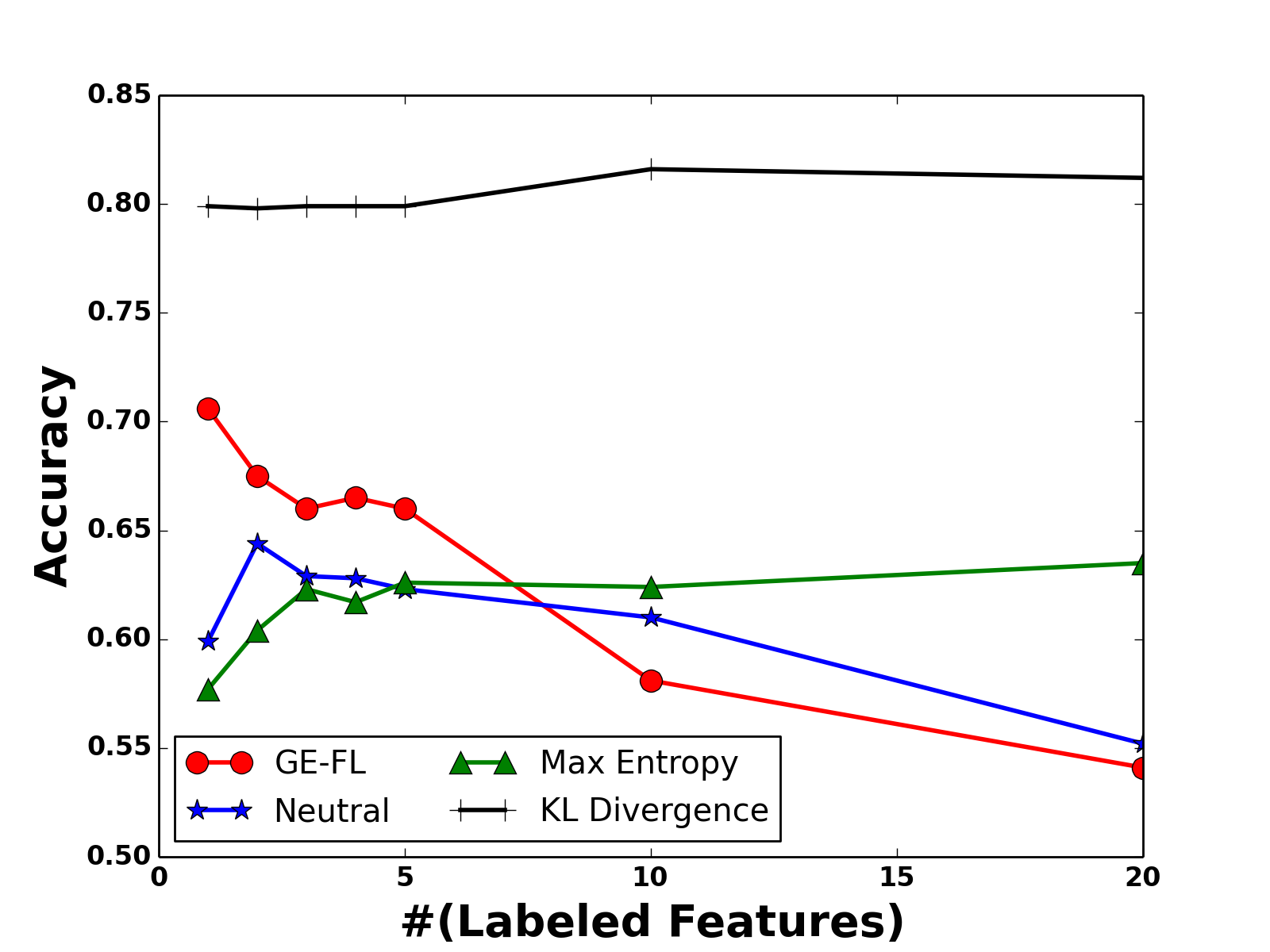}
	}
	\caption{Performance with unbalanced labeled features, tested on the {\em movie} dataset. Randomly select from the feature pool $t$ (x-axis) labeled features for one class, and select $1$ feature for the other. The unbalanced datasets in (b) are constructed by randomly removing 75\% of the {\em positive} documents.} \label{fig-ulf}
\end{figure*}

\begin{figure*}[!ht]
\setcounter{subfigure}{0}
\centering
	\subfigure[balanced dataset] {
		
		\includegraphics[width=190px]{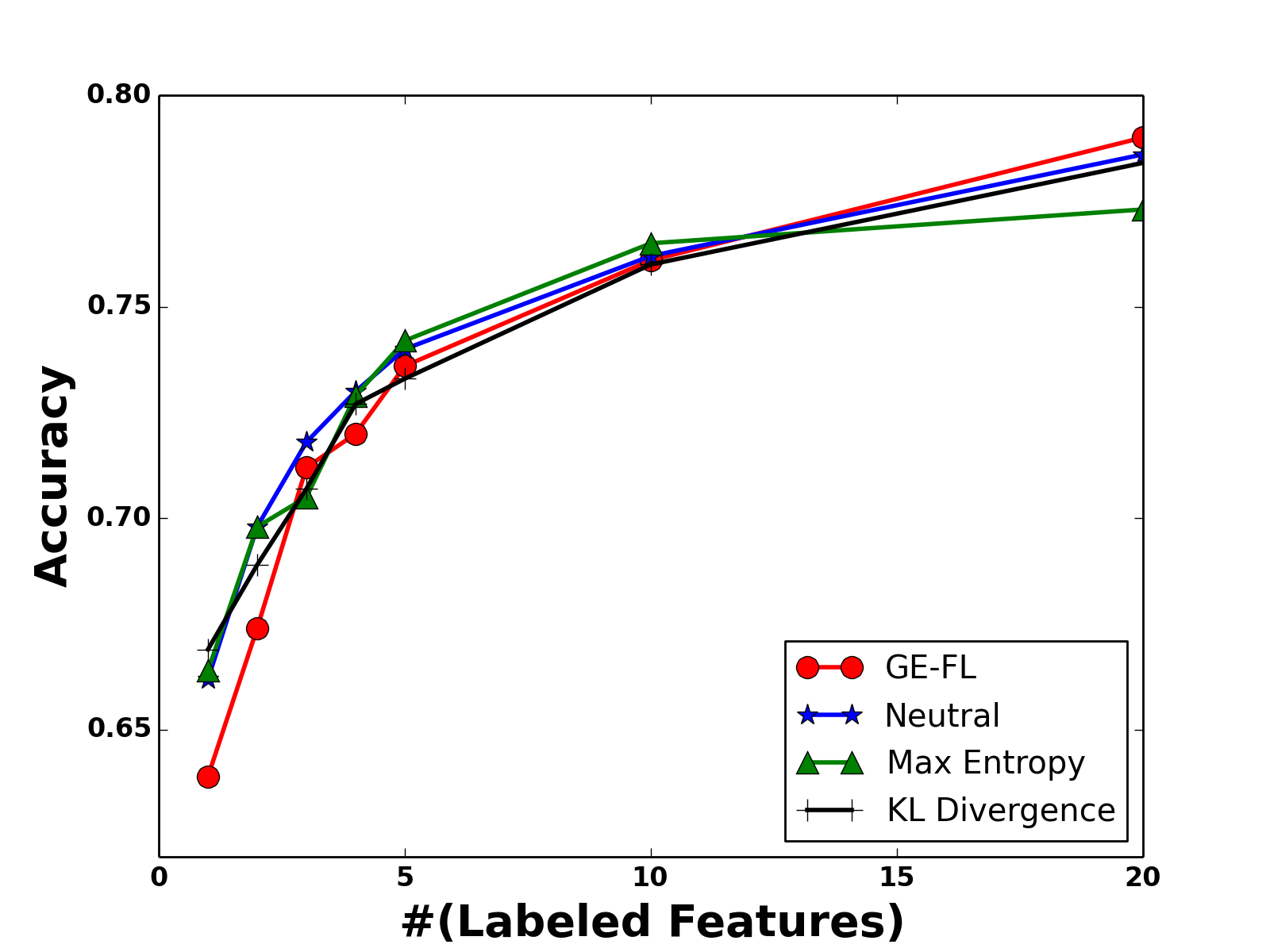}
	}
	\subfigure[unbalanced dataset(1:4)] {
		\includegraphics[width=190px]{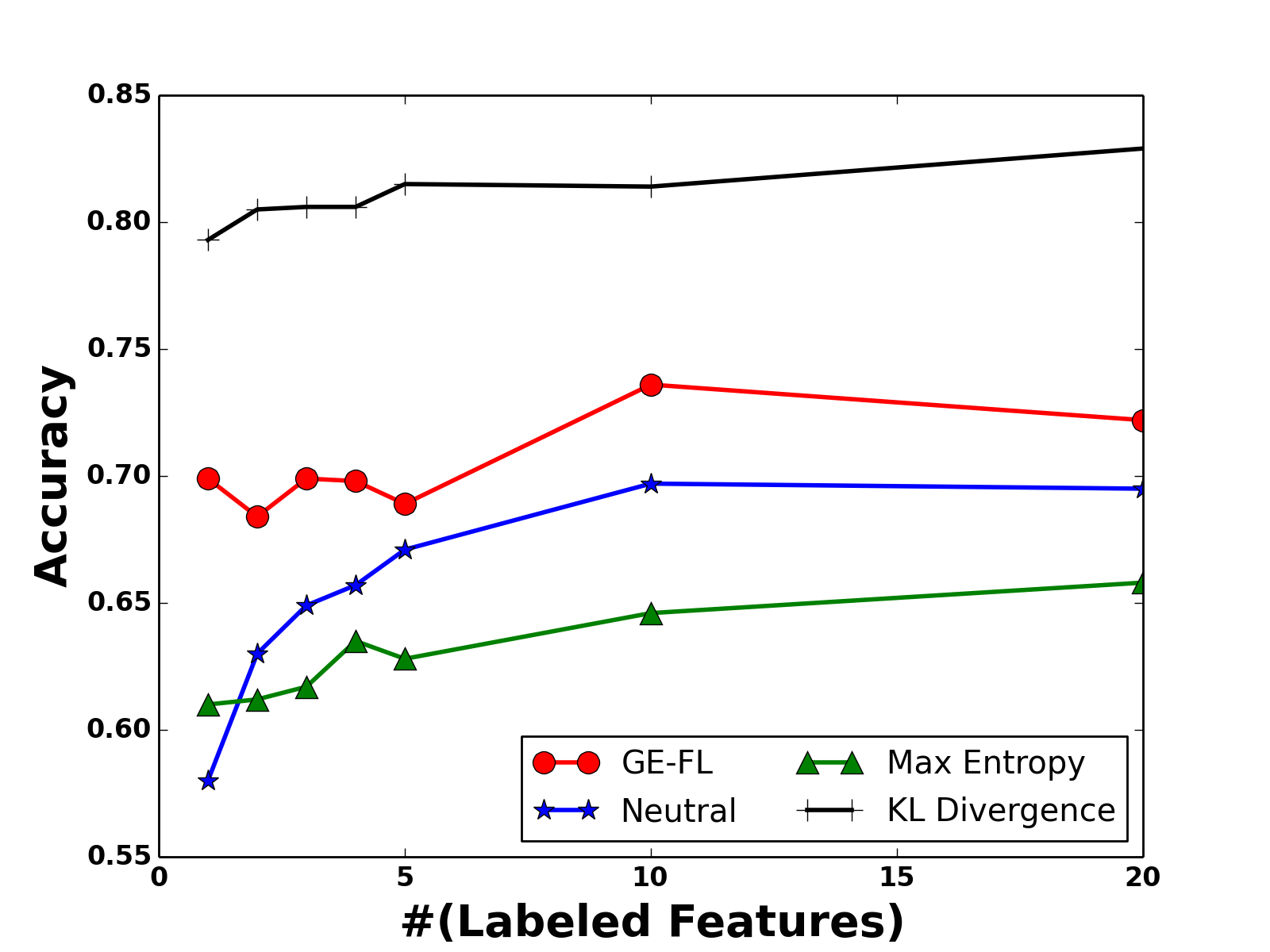}
	}
	\caption{Performance with balanced labeled features, tested on the {\em movie} dataset. Randomly select from the feature pool $t$ (x-axis) labeled features for each class. The unbalanced datasets in (b) are constructed by randomly removing 75\% of the {\em positive} documents.} \label{fig-blf}
\end{figure*}

\subsection{With Unbalanced Labeled Features}
\label{sulf}

In this section, we evaluate our approach when there is unbalanced knowledge on the categories to be classified. The labeled features are obtained through information gain.
Two settings are chosen:

\textbf{(a)} We randomly select $t \in [1, 20]$ features from the feature pool for one class, and only one feature for the other. The original balanced {\em movie} dataset is used (positive:negative=1:1).

\textbf{(b)} Similar to (a), but the dataset is unbalanced, obtained by randomly removing 75\% {\em positive} documents (positive:negative=1:4).

\begin{figure*}[t]
\centering
\setcounter{subfigure}{0}
	\subfigure[balanced labeled features] {
		
		\includegraphics[width=190px]{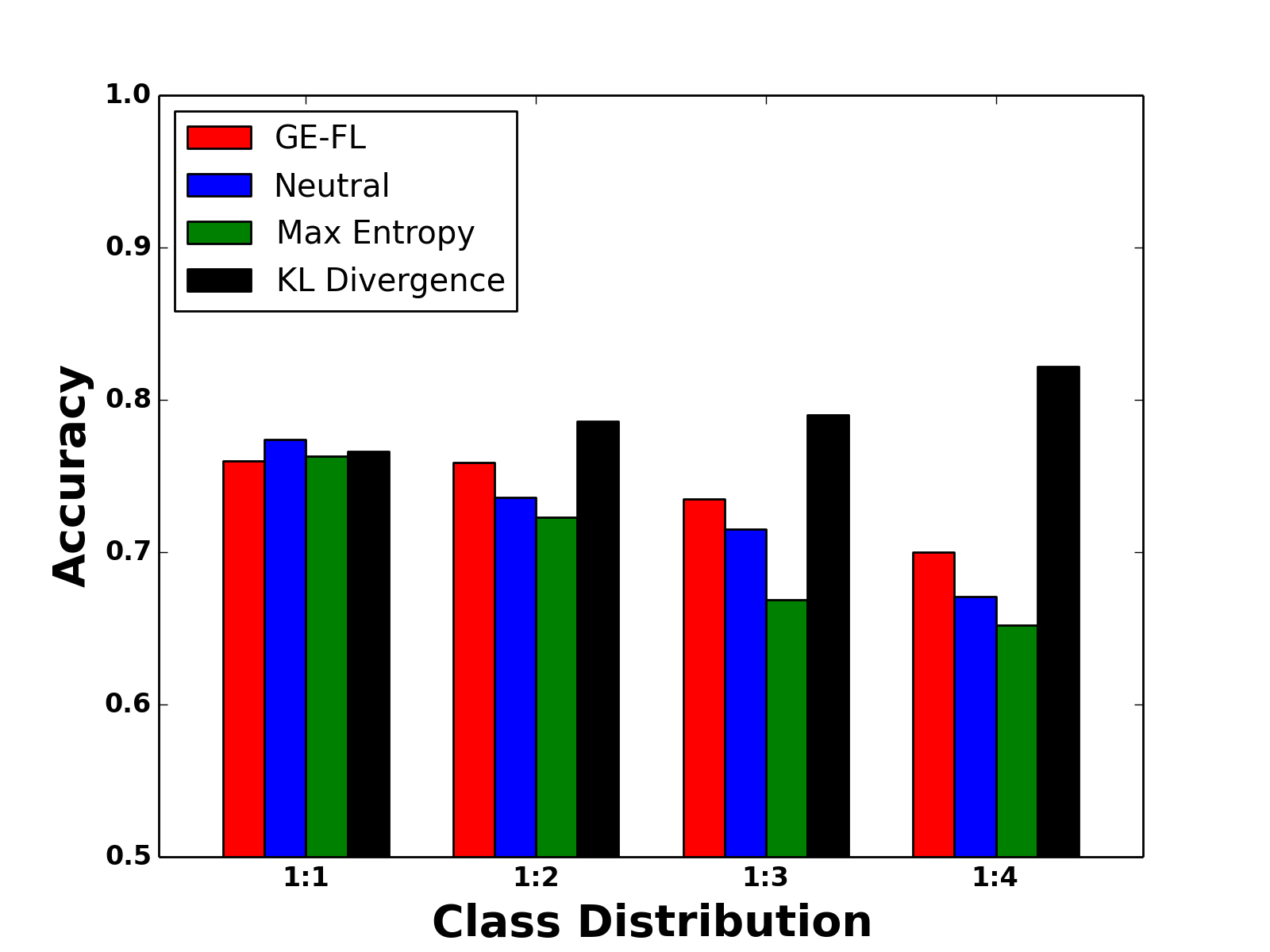}
	}
	\subfigure[unbalanced labeled features] {
		\includegraphics[width=190px]{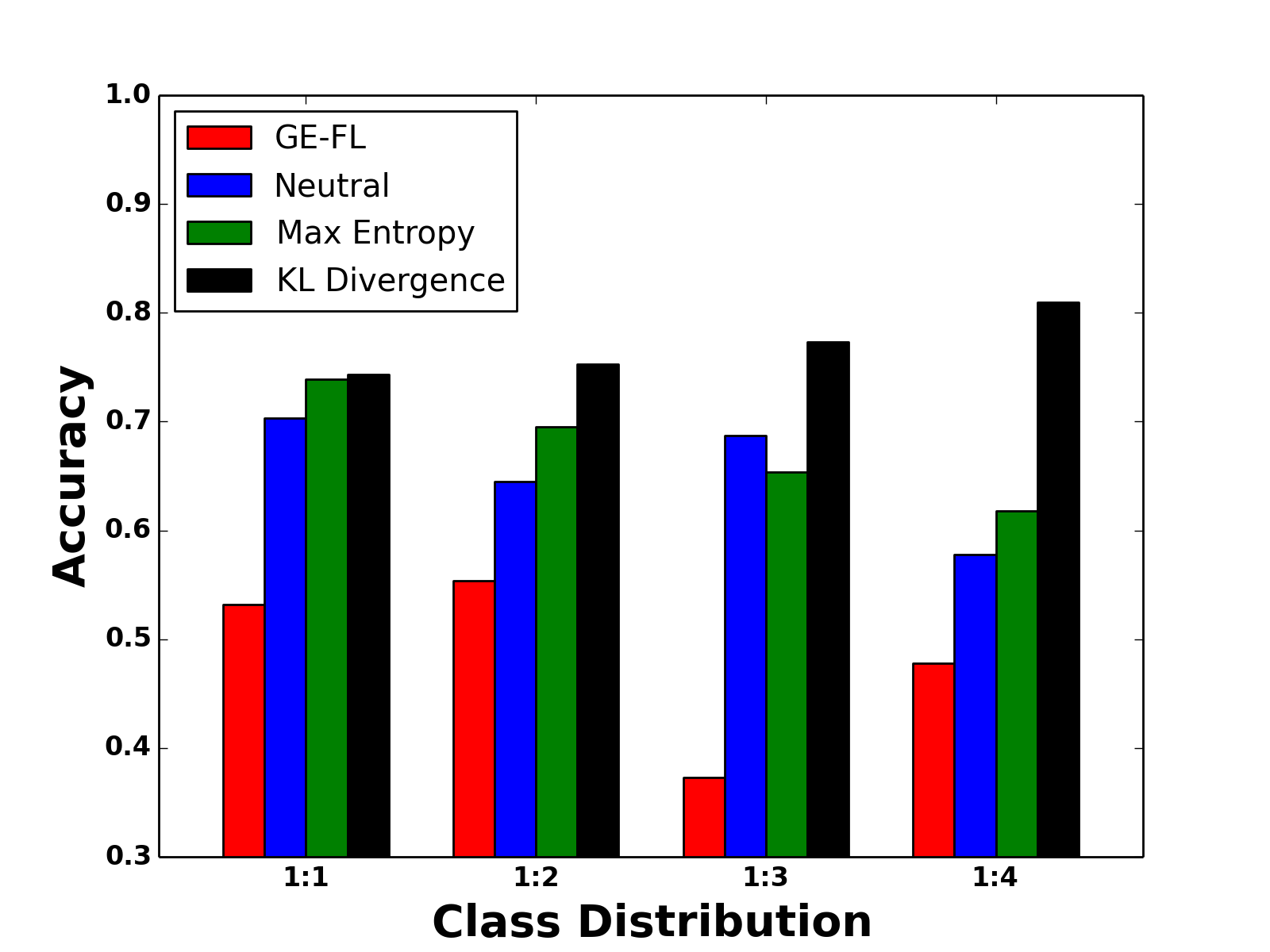}
	}
	\caption{Influence of unbalanced class distribution, tested on the {\em movie} dataset. We provide 10 labeled features for each class in (a), and, 10 for one class and 1 for the other in (b). The three unbalanced datasets are constructed by randomly removing 50\%, 67\%, 75\% documents of the {\em positive} class.} \label{fig-iud}

\end{figure*}



As shown in Figure \ref{fig-ulf}, Maximum entropy principle shows improvement only on the balanced case. An obvious reason is that maximum entropy only favors uniform distribution.

Incorporating Neutral features performs similarly to maximum entropy since we assume that neutral words are uniformly distributed. Its accuracy decreases slowly when the number of labeled features becomes larger ($t>4$) (Figure \ref{fig-ulf}(a)), suggesting that the model gradually biases to the class with more labeled features, just like GE-FL. 


Incorporating the KL divergence of class distribution performs much better than GE-FL on both balanced and unbalanced datasets. This shows that it is effective to control the unbalance in labeled features and in the dataset.

\subsection{With Balanced Labeled Features}
\label{sblf}
We also compare with the baseline when the labeled features are balanced.
Similar to the experiment above, the labeled features are obtained by information gain.
Two settings are experimented with:

\textbf{(a)} We randomly select $t \in [1, 20]$ features from the feature pool for each class, and conduct comparisons on the original balanced {\em movie} dataset (positive:negtive=1:1).

\textbf{(b)} Similar to (a), but the class distribution is unbalanced, by randomly removing 75\% {\em positive} documents (positive:negative=1:4).

Results are shown in Figure \ref{fig-blf}. When the dataset is balanced (Figure \ref{fig-blf}(a)), there is little difference between GE-FL and our methods. The reason is that the proposed regularization terms provide no additional knowledge to the model and there is no bias in the labeled features.
On the unbalanced dataset (Figure \ref{fig-blf}(b)), incorporating KL divergence is much better than GE-FL since we provide additional knowledge(the true class distribution), but maximum entropy and neutral features are much worse because forcing the model to approach the uniform distribution misleads it.

\subsection{With Unbalanced Class Distributions}
\label{sud}

\begin{figure*}[!ht]
\setcounter{subfigure}{0}
\centering
	\subfigure[balanced dataset] {
		
		\includegraphics[width=190px]{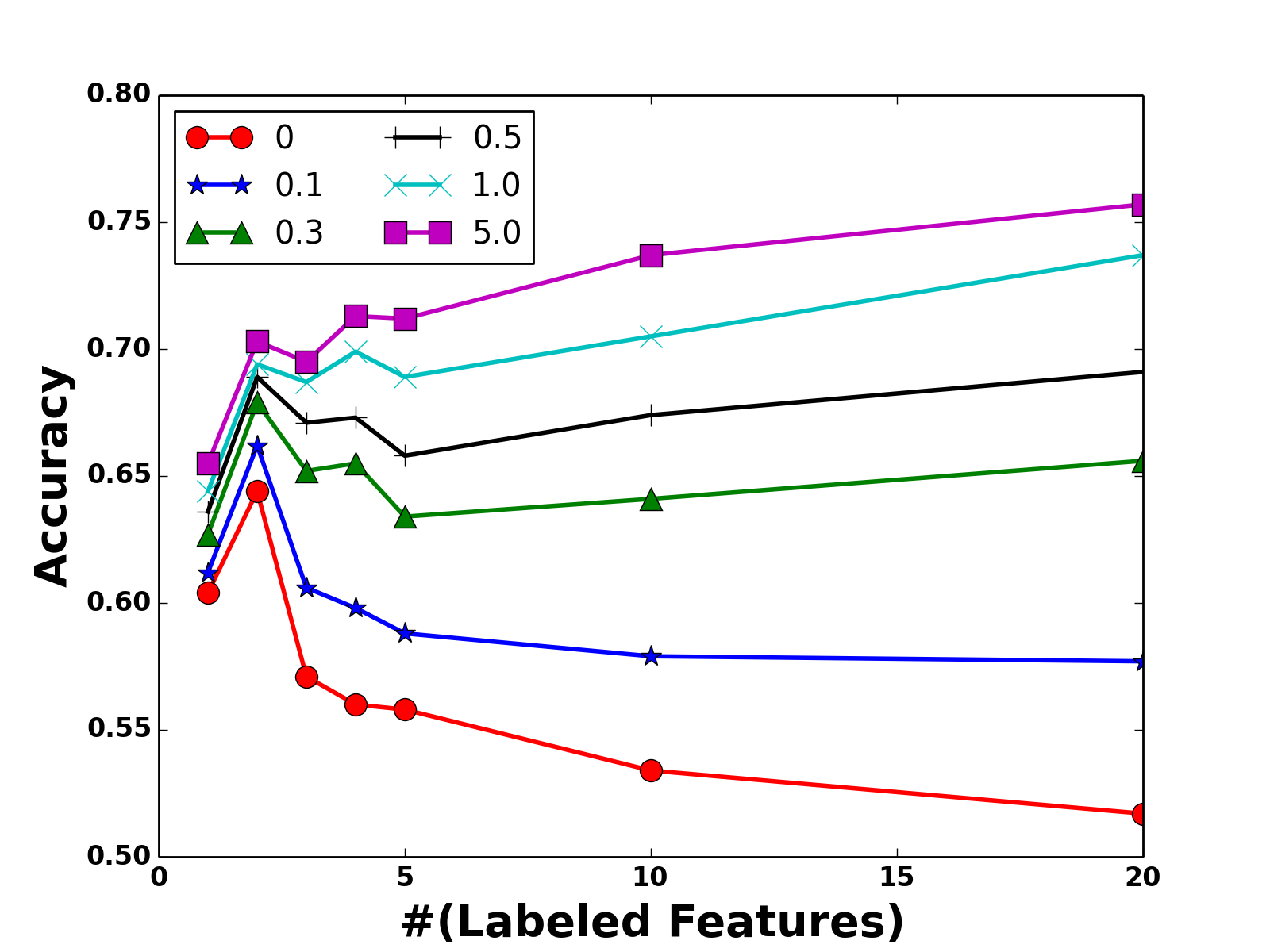}
	}
	\subfigure[unbalanced dataset(1:4)] {
		\includegraphics[width=190px]{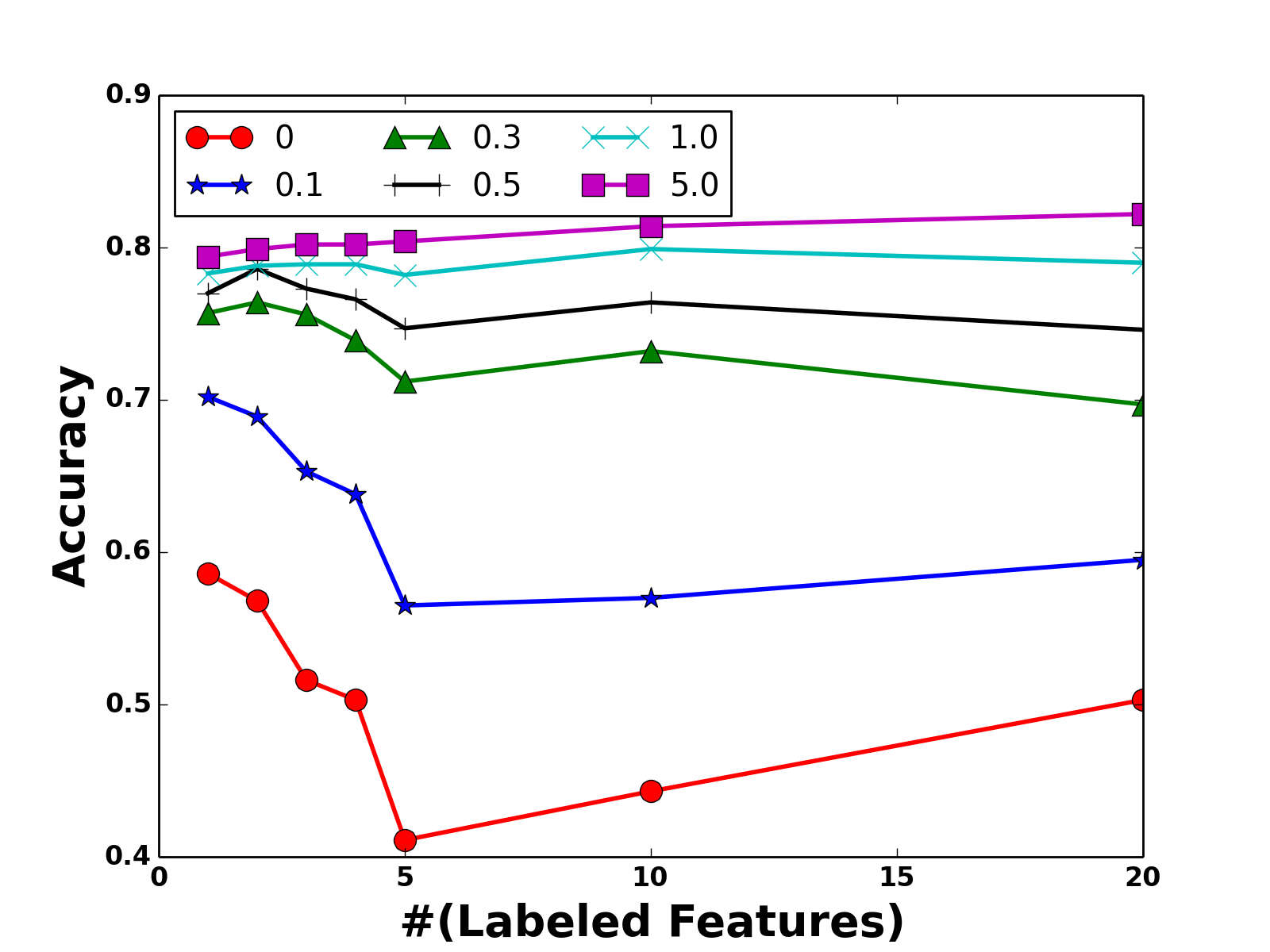}
	}
	\caption{The influence of $\lambda$, tested on the {\em movie} dataset. Randomly select from the feature pool $t$ (x-axis) labeled features for one class, and keep $1$ feature for the other. The unbalanced datasets in (b) are constructed by randomly removing 75\% documents of the {\em positive} class.} \label{fig-il}
\end{figure*}

Our methods are also evaluated on  datasets with different unbalanced class distributions. We manually construct several {\em movie} datasets with class distributions of 1:2, 1:3, 1:4 by randomly removing 50\%, 67\%, 75\% {\em positive} documents. The original balanced {\em movie} dataset is used as a control group.
We test with both balanced and unbalanced labeled features.
For the balanced case, we randomly select 10 features from the feature pool for each class, and for the unbalanced case, we select 10 features for one class, and 1 feature for the other. Results are shown in Figure \ref{fig-iud}.

Figure \ref{fig-iud} (a) shows that when the dataset and the labeled features are both balanced, there is little difference between our methods and GE-FL(also see Figure \ref{fig-blf}(a)). But when the class distribution becomes more unbalanced, the difference becomes more remarkable. Performance of neutral features and maximum entropy decrease significantly but incorporating KL divergence increases remarkably. This suggests if we have more accurate knowledge about class distribution, KL divergence can guide the model to the right direction.

Figure \ref{fig-iud} (b) shows that when the labeled features are unbalanced, our methods significantly outperforms GE-FL.
Incorporating KL divergence is robust enough to control unbalance both in the dataset and in labeled features while the other three methods are not so competitive.

\subsection{The Influence of $\lambda$}
\label{sil}

We present the influence of $\lambda$ on the method that incorporates KL divergence in this section. Since we simply set $\lambda = \beta |K|$, we just tune $\beta$ here. Note that when $\beta = 0$, the newly introduced regularization term is disappeared, and thus the model is actually GE-FL. Again, we test the method with different $\lambda$ in two settings:

\begin{table*}[!htp]
\begin{minipage}{\textwidth}
\centering
\begin{tabular}{|c|c|c|c|c|}
\hline
Dataset&GE-FL&Neutral Features&Max Entropy&KL Divergence\\
\hline
movie&0.623&\bf{0.672}&\bf{0.681}&\bf{0.684} \\
\hline
sraa&0.559&\bf{0.628}&\bf{0.618}&0.547 \\
\hline
webkb&0.615&\bf{0.685}&\bf{0.640}&\bf{0.646} \\
\hline
med-space&0.927&0.921&\bf{0.936}&\bf{0.928} \\
ibm-mac&0.817&0.796&\bf{0.837}&\bf{0.833} \\
baseball-hockey&0.915&\bf{0.935}&\bf{0.925}&\bf{0.923} \\
20 newsgroups&0.667&\bf{0.669}&\bf{0.680}&\bf{0.677} \\
\hline
financial-healthcare&0.588&\bf{0.618}&\bf{0.618}&0.507 \\
sector.top&0.596&\bf{0.653}&0.581&\bf{0.639} \\
\hline
\end{tabular}
\caption{Performance using LDA-features. Bold text means the method performs better than GE-FL.} \label{tab-rd}
\end{minipage}
\end{table*}

\textbf{(a)} We randomly select $t \in [1, 20]$ features from the feature pool for one class, and only one feature for the other class. The original balanced {\em movie} dataset is used (positive:negative=1:1).

\textbf{(b)} Similar to (a), but the dataset is unbalanced, obtained by randomly removing 75\% {\em positive} documents (positive:negative=1:4).

Results are shown in Figure \ref{fig-il}. As expected, $\lambda$ reflects how strong the regularization is. The model tends to be closer to our preferences with the increasing of $\lambda$ on both cases.

\subsection{Using LDA Selected Features}
\label{sfud}

We compare our methods with GE-FL on all the 9 datasets in this section. Instead of using features obtained by information gain, we use LDA to select labeled features. Unlike information gain, LDA does not employ any instance labels to find labeled features. In this setting, we can build classification models without any instance annotation, but just with labeled features.

Table \ref{tab-rd} shows that our three methods significantly outperform GE-FL. Incorporating neutral features performs better than GE-FL on 7 of the 9 datasets, maximum entropy is better on 8 datasets, and KL divergence better on 7 datasets.

LDA selects out the most predictive features as labeled features without considering the balance among classes. GE-FL does not exert any control on such an issue, so the performance is severely suffered. Our methods introduce auxiliary regularization terms to control such a bias problem and thus promote the model significantly.

\section{Related Work}
There have been much work that incorporate prior knowledge into learning, and two related lines are surveyed here. One is to use prior knowledge to label unlabeled instances and then apply a standard learning algorithm. The other is to constrain the model directly with prior knowledge.

Liu et al.\shortcite{text} manually labeled features which are highly predictive to unsupervised clustering assignments and use them to label unlabeled data. Chang et al.\shortcite{guiding} proposed constraint driven learning. They first used constraints and the learned model to annotate unlabeled instances, and then updated the model with the newly labeled data. Daum{\'e} \shortcite{daume2008cross} proposed a self training method in which several models are trained on the same dataset, and only unlabeled instances that satisfy the cross task knowledge constraints are used in the self training process.
 
MaCallum et al.\shortcite{gec} proposed generalized expectation(GE) criteria which formalised the knowledge as constraint terms about the expectation of the model into the objective function.Gra{\c{c}}a et al.\shortcite{pr} proposed posterior regularization(PR) framework which projects the model's posterior onto a set of distributions that satisfy the auxiliary constraints. Druck et al.\shortcite{ge-fl} explored constraints of labeled features in the framework of GE by forcing the model's predicted feature distribution to approach the reference distribution. Andrzejewski et al.\shortcite{andrzejewski2011framework} proposed a framework in which general domain knowledge can be easily incorporated into LDA. Altendorf et al.\shortcite{altendorf2012learning} explored monotonicity constraints to improve the accuracy while learning from sparse data. Chen et al.\shortcite{chen2013leveraging} tried to learn comprehensible topic models by leveraging multi-domain knowledge.

Mann and McCallum \shortcite{simple,generalized} incorporated not only labeled features but also other knowledge like class distribution into the objective function of GE-FL. But they discussed only from the semi-supervised perspective and did not investigate into the robustness problem, unlike what we addressed in this paper.

There are also some active learning methods trying to use prior knowledge. Raghavan et al.\shortcite{feedback} proposed to use feedback on instances and features interlacedly, and demonstrated that feedback on features boosts the model much. Druck et al.\shortcite{active} proposed an active learning method which solicits labels on features rather than on instances and then used GE-FL to train the model. 

\section{Conclusion and Discussions}

This paper investigates into the problem of how to leverage prior knowledge robustly in learning models. We propose three regularization terms on top of generalized expectation criteria. As demonstrated by the experimental results, the performance can be considerably improved when taking into account these factors. Comparative results show that our proposed methods is more effective and works more robustly against baselines. To the best of our knowledge, 
this is the first work to address the robustness problem of leveraging knowledge, and may inspire other research.

We then present more detailed discussions about the three regularization methods. Incorporating neutral features is the simplest way of regularization, which doesn't require any modification of GE-FL but just finding out some common features. But as Figure \ref{fig-ulf}(a) shows, only using neutral features are not strong enough to handle extremely unbalanced labeled features.

The maximum entropy regularization term shows the strong ability of controlling unbalance.

This method doesn't need any extra knowledge, and is thus suitable when we know nothing about the corpus. But this method assumes that the categories are uniformly distributed, which may not be the case in practice, and it will have a degraded performance if the assumption is violated (see Figure \ref{fig-ulf}(b), Figure \ref{fig-blf}(b), Figure \ref{fig-iud}(a)). 

The KL divergence performs much better on unbalanced corpora than other methods. The reason is that KL divergence utilizes the reference class distribution and doesn't make any assumptions. The fact suggests that additional knowledge does benefit the model.

However, the KL divergence term requires providing the true class distribution.
Sometimes, we may have the exact knowledge about the true distribution, but sometimes we may not.
Fortunately, the model is insensitive to the true distribution and therefore a rough estimation of the true distribution is sufficient. In our experiments, when the true class distribution is 1:2, where
the reference class distribution is set to 1:1.5/1:2/1:2.5, the
accuracy is 0.755/0.756/0.760 respectively. This provides us the possibility to perform simple computing on the corpus to obtain the distribution in reality. Or, we can set the distribution roughly with domain expertise.

\clearpage
\bibliographystyle{naaclhlt2015.bst}
\bibliography{naacl}

\begin{thebibliography}{}

\bibitem[\protect\citename{Altendorf \bgroup et al.\egroup
  }2012]{altendorf2012learning}
Eric~E Altendorf, Angelo~C Restificar, and Thomas~G Dietterich.
\newblock 2012.
\newblock Learning from sparse data by exploiting monotonicity constraints.
\newblock {\em arXiv preprint arXiv:1207.1364}.

\bibitem[\protect\citename{Andrzejewski and Zhu}2009]{andrzejewski2009latent}
David Andrzejewski and Xiaojin Zhu.
\newblock 2009.
\newblock Latent dirichlet allocation with topic-in-set knowledge.
\newblock In {\em Proceedings of the NAACL HLT 2009 Workshop on Semi-Supervised
  Learning for Natural Language Processing}, pages 43--48. Association for
  Computational Linguistics.

\bibitem[\protect\citename{Andrzejewski \bgroup et al.\egroup
  }2011]{andrzejewski2011framework}
David Andrzejewski, Xiaojin Zhu, Mark Craven, and Benjamin Recht.
\newblock 2011.
\newblock A framework for incorporating general domain knowledge into latent
  dirichlet allocation using first-order logic.
\newblock In {\em IJCAI Proceedings-International Joint Conference on
  Artificial Intelligence}, volume~22, page 1171.

\bibitem[\protect\citename{Blei \bgroup et al.\egroup }2003]{lda}
David~M Blei, Andrew~Y Ng, and Michael~I Jordan.
\newblock 2003.
\newblock Latent dirichlet allocation.
\newblock {\em the Journal of machine Learning research}, 3:993--1022.

\bibitem[\protect\citename{Chang \bgroup et al.\egroup }2007]{guiding}
Ming-Wei Chang, Lev Ratinov, and Dan Roth.
\newblock 2007.
\newblock Guiding semi-supervision with constraint-driven learning.
\newblock In {\em ANNUAL MEETING-ASSOCIATION FOR COMPUTATIONAL LINGUISTICS},
  volume~45, page 280. Citeseer.

\bibitem[\protect\citename{Chen \bgroup et al.\egroup
  }2013]{chen2013leveraging}
Zhiyuan Chen, Arjun Mukherjee, Bing Liu, Meichun Hsu, Malu Castellanos, and
  Riddhiman Ghosh.
\newblock 2013.
\newblock Leveraging multi-domain prior knowledge in topic models.
\newblock In {\em Proceedings of the Twenty-Third international joint
  conference on Artificial Intelligence}, pages 2071--2077. AAAI Press.

\bibitem[\protect\citename{Daum{\'e}~III}2008]{daume2008cross}
Hal Daum{\'e}~III.
\newblock 2008.
\newblock Cross-task knowledge-constrained self training.
\newblock In {\em Proceedings of the conference on empirical methods in natural
  language processing}, pages 680--688. Association for Computational
  Linguistics.

\bibitem[\protect\citename{Druck \bgroup et al.\egroup }2008]{ge-fl}
Gregory Druck, Gideon Mann, and Andrew McCallum.
\newblock 2008.
\newblock Learning from labeled features using generalized expectation
  criteria.
\newblock In {\em Proceedings of the 31st annual international ACM SIGIR
  conference on Research and development in information retrieval}, pages
  595--602. ACM.

\bibitem[\protect\citename{Druck \bgroup et al.\egroup }2009]{active}
Gregory Druck, Burr Settles, and Andrew McCallum.
\newblock 2009.
\newblock Active learning by labeling features.
\newblock In {\em Proceedings of the 2009 Conference on Empirical Methods in
  Natural Language Processing: Volume 1-Volume 1}, pages 81--90. Association
  for Computational Linguistics.

\bibitem[\protect\citename{Gra{\c{c}}a \bgroup et al.\egroup }2007]{pr}
Joao~V Gra{\c{c}}a, Kuzman Ganchev, and Ben Taskar.
\newblock 2007.
\newblock Expectation maximization and posterior constraints.

\bibitem[\protect\citename{Haghighi and Klein}2006]{haghighi2006prototype}
Aria Haghighi and Dan Klein.
\newblock 2006.
\newblock Prototype-driven learning for sequence models.
\newblock In {\em Proceedings of the main conference on Human Language
  Technology Conference of the North American Chapter of the Association of
  Computational Linguistics}, pages 320--327. Association for Computational
  Linguistics.

\bibitem[\protect\citename{Li \bgroup et al.\egroup }2010]{li2010sentiment}
Fangtao Li, Minlie Huang, and Xiaoyan Zhu.
\newblock 2010.
\newblock Sentiment analysis with global topics and local dependency.
\newblock In {\em AAAI}.

\bibitem[\protect\citename{Liu \bgroup et al.\egroup }2004]{text}
Bing Liu, Xiaoli Li, Wee~Sun Lee, and Philip~S Yu.
\newblock 2004.
\newblock Text classification by labeling words.
\newblock In {\em AAAI}, volume~4, pages 425--430.

\bibitem[\protect\citename{Mann and McCallum}2007]{simple}
Gideon~S Mann and Andrew McCallum.
\newblock 2007.
\newblock Simple, robust, scalable semi-supervised learning via expectation
  regularization.
\newblock In {\em Proceedings of the 24th international conference on Machine
  learning}, pages 593--600. ACM.

\bibitem[\protect\citename{Mann and McCallum}2010]{generalized}
Gideon~S Mann and Andrew McCallum.
\newblock 2010.
\newblock Generalized expectation criteria for semi-supervised learning with
  weakly labeled data.
\newblock {\em The Journal of Machine Learning Research}, 11:955--984.

\bibitem[\protect\citename{McCallum \bgroup et al.\egroup }2007]{gec}
Andrew McCallum, Gideon Mann, and Gregory Druck.
\newblock 2007.
\newblock Generalized expectation criteria.
\newblock {\em Computer science technical note, University of Massachusetts,
  Amherst, MA}.

\bibitem[\protect\citename{Raghavan and Allan}2007]{raghavan2007interactive}
Hema Raghavan and James Allan.
\newblock 2007.
\newblock An interactive algorithm for asking and incorporating feature
  feedback into support vector machines.
\newblock In {\em Proceedings of the 30th annual international ACM SIGIR
  conference on Research and development in information retrieval}, pages
  79--86. ACM.

\bibitem[\protect\citename{Raghavan \bgroup et al.\egroup }2006]{feedback}
Hema Raghavan, Omid Madani, and Rosie Jones.
\newblock 2006.
\newblock Active learning with feedback on features and instances.
\newblock {\em The Journal of Machine Learning Research}, 7:1655--1686.

\bibitem[\protect\citename{Schapire \bgroup et al.\egroup }2002]{boosting}
Robert~E Schapire, Marie Rochery, Mazin Rahim, and Narendra Gupta.
\newblock 2002.
\newblock Incorporating prior knowledge into boosting.
\newblock In {\em ICML}, volume~2, pages 538--545.

\end{thebibliography}

\end{document}